# An Innovative Human-Robot Cognitive Coupling in Industrial Robot Control and Manufacturing

Yusuke Inoune, and Agunar Prayit[1]

*Abstract*—this paper demonstrates the groundwork for the structure and nature of Human-Robot Cognitive Coupling. The human mind is best at associating objects, while digital devices can only compare. Successful communication between robot and human agents involves translating human thought into unique numbers that can be compared, and translating these numbers back into a form that humans can understand. A cognitive robot is one capable of interpreting its own operational data. Working from a D&D project standpoint with a telerobot, a system is designed to utilize sensor data, extract usable profiles and information from them, and then apply them as a means of self-testing for nominal/off-nominal operations. If operations are considered off-nominal, the telerobot begins a sequence of recalling and comparing previous off-nominal occurrences to determine potential corrections. In this manner, a telerobot can apply previous 'experiences' in order to learn without the need for operator input. In the cognitive telerobotics, knowledge base is an important component to store required information. In this paper, a tool-centric knowledge base with learning capability is presented. In this paper, the knowledge can be updated based on the reinforcement learning algorithm at both global and local levels and knowledge fusion by coupling with operator's cognition.

## I. INTRODUCTION

Robot intelligence began as a topic for science fiction in the 1940s, [1]. Since then, the robotics community has been working diligently at trying to make these ideas a reality by either designing a wide variety of robots i.e. AGV, UAV, robotic arm, humanoid [2], [3] or automating and improving vehicles like two-wheel vehicles i.e. bicycle, motorcycle and hover-board [4], [5]. The term Artificial Intelligence was coined in the 1970s [6] as research began realizing on software advances needed to create adaptable robots. As the 1990s ended, many in the robotics community were surprised at how little progress has been made in the development of truly intelligent robots. Specifically, there is still a need for robots that can work with humans in unknown and unstructured environments. Strong initiatives from the European Union have recently increased the research in this area [7]. This new work is being called Cognitive Robotics and focuses on increasing the ability of a robot to reason.

This reasoning ability can be achieved from two fronts. One direction is focusing on the ability of a robot to reason on its own. This is achieved by giving the robot quantitative models to compare expected or desired situations with its present situation. The emphasis is on how machine learning techniques are affected by system and tool models. The other direction is called cognitive coupling. This approach focuses on the interaction of a human with a robot. The robot reasons by leaning on the reasoning ability of a human operator. While a truly successful cognitive robot must excel in both self-reasoning and cognitive coupling, this paper focuses on cognitive coupling. Specifically, we describe a framework in which a cognitive connection can be established between a human and a robot, as shown in Fig. 1.

When humans work in unstructured and unknown environments, we use three basic tools to make sense of the world around us:

1) Memory (recalling past experiences)
2) Sensors (understanding current conditions)
3) Cognitive coupling (communicating with and learning from others)

Human cognitive coupling is two-way communication on a reasoning level. Consider an apprentice craftsman relationship as an example. The apprentice explains what he is trying to do, the master taps into her knowledge base and explains how she has accomplished this task, and the apprentice adds that knowledge to his own. The apprentice now has a better understanding of the task. When a problem arises, the apprentice explains what is occurring and the master suggests possible causes and solutions. If the communication is successful, the apprentice can fix the problem and learn how to fix these types of problems in the future.

The goal of robotic cognitive coupling is to allow a robot to assume the role of an apprentice. The concept of a generalized cognitive agent and cognitive coupling is illustrated in Fig. 1.

1) Less human fatigue the human has assumed the role of supervisor instead of operator.
2) Less human training the human will be communicating with the robot in the same manner s/he communicates with people.
3) Cognitive coupling (communicating with and learning from others)

The first objective of this paper is to lay a groundwork for the structure and nature of this Human-Robot Cognitive Coupling. At the heart of this coupling is the realization of the nature of human and computer thought. The human mind is best at associating objects, while digital devices can only compare. Successful communication between robot and human involves translating human thought into unique numbers that can be compared, and translating these numbers back into a form that humans can understand.

[1]Yusuke Inoune and agunar prayit are with Electrical Engineering Department at Sriwijaya University, Palembang, South Sumatra, Indonesia agunarprayit1992@gmail.com

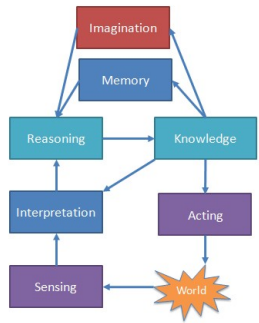

Fig. 1. Cognitive agent and cognitive coupling.

A cognitive robot is often defined as one equipped "with planning, reasoning, navigation, manipulation and other related skills necessary to interact with and operate in the non-social environment" [8]. Sensors and manipulator capabilities allow for the direct interaction between the robot and its environment.

Initial attempts at this approach resulted in what is known as Behavior-based Robotics (BBR). In this method of , the robot uses preprogrammed responses to specified stimuli. Another approach is to program specific parts of a repetitive sequence (generating an operator-based telerobot) as demonstrated by Noakes [9] such that the operator is still in overall control while the robot performs the specified task sequences. This method allows for a broad adaptability, but still relies heavily on a human operator to judge the final course of actions to be taken.

Another interesting approach to developing a cognitive system came from manufacturing [10], [11], [12], [13]. Rather than create a singular cognitive approach specific to a setting, *Goc* and *Gaeta* designed a cognitive approach that would adapt first to its environment and then begin working as a corrective agent [14]. This means allowed for a general "all-purpose" agent to be sent out that could then be customized to its working environment (such as a factory line/floor). Although this provides more in the means of robot-based judgment, this still relies on a fairly stable, known environment. Therefore, the second objective of this project is to transform the data from the sensors into a useable format and then it is interpreted.

A telerobot is a remotely operated robot that can accept instructions from a distance. Usually there are a master robot and a slave robot connected with each other through communication network. Simple examples of master robots are: joy sticks, haptic devices, mouses and so on. Slave robot is designed to facilitate their tasks: large manipulators are used in radioactive cleanup task while small robots are widely used in minimal invasive surgeries. Telerobots have to use a wide range of tools due to the complexity of their task. Most of the time, the method of using these tool is limited to hard coded programs that does not evolve at all[15].

Humans learn by experiences. Experienced workers are better paid because they do their job faster and better, know their tasks and tools better. The advantage of migrating this learning capability to telerobots is obvious. If a robot can learn their workspace and other conditions by itself, both the work load of the operators and the cost could be reduced [16] .

To enable the robot to benefit from the experience and learn new skills, knowledge modeling and processing become the key components to make the knowledge comprehensive for the computer which only accept and process information in the form of numbers [17], [18], [19]. The tasks, especially complex ones, usually require many different kinds of knowledge, that infers the specific actions stored in the knowledge base. The robot needs to know which objects are involved, where are objects, which tool is used, how to use the tool and so forth.

The third objective of this project is to develop a dynamic tooling knowledge base for the knowledge modeling and processing. There are several challenges. Firstly, all knowledge has to be convert into numbers to be evaluated and analysed. Raw data are usually in large amounts and storing them can be very space consuming. Then, the environment is dynamic, and most of the information in the knowledge base is obtained using the sensor with a certain degree of uncertainty. It is difficult to eliminate these contradictions using only deterministic representations in the whole processing of task. Thus, how to continually update the knowledge to keep it consistent with the state of the world is one of great challenges. The last challenge is the unspecific task from the operator's command, which does not include all required information to perform the task. In this project, a tooling knowledge base with learning capability was proposed. There were two learning modes to enhance the capability of robot. The first mode is learning from experience based on a reinforcement learning algorithm, followed by the second mode in which robot can learn new skill from the operator by incorporating the existing knowledge.

The rest of this paper is organized as follows. Section II will discuss human and robot communication for cognitive coupling. Section III describes transforming the data from the sensors into a useable format and then it is interpreted. Section IV explains how to develop a dynamic tooling knowledge base for the knowledge modeling and processing.

## II. Human Robot Communication for Cognitive Coupling

### A. Human to Robot Communication

The dominant way humans communicate with other humans is through voice. Humans are very adept to using spoken language to provide cognitive coupling to each other. Human speech is a multiplex signal. The pattern of the words and sentences, the absolute pitch, the relative pitch of the words within a sentence, the speed of the sentence, the absolute and relative volume of the sentence, and the tone of the sentence all provide information about who is speaking and what is being said. An entry-level cognitive robot might not need to process all of these signals, but each of these channels of communication carry meaning.

In addition to voice, humans also communicate with body language. This is a very complex and subtle way of suggesting meaning, and many times we communicate with body language at a level below human consciousness. For an entry-level cognitive robot, this layer of communication should not

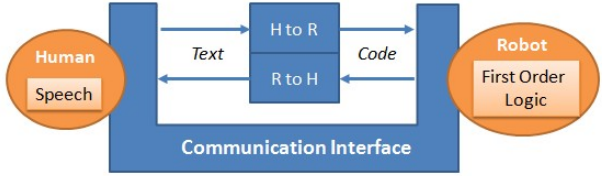

(a) Human and robot cognitive coupling layer.

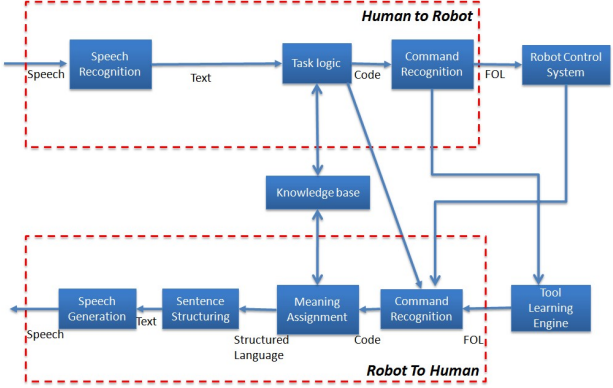

(b) Cognitive coupling detailed diagram.

Fig. 2. Human and robot cognitive coupling layer

be considered. The critical task of a cognitive robot is to understand what the human is trying to communicate, not to attempt to understand what the human is not trying to communicate.

*1) Speech Recognition:* The first step to the communication process is to understand what words are being spoken. This area has been heavily researched and numerous available commercial products exist to translate the spoken word into written text. For example, Apple uses Siri, Google uses Google Voice, Microsofts program is called Speech Recognition. All of these products work in a similar manner. The program hears the sounds, parses the sounds into words, assigns meanings to the words and executes various tasks based on the human request.

The task of parsing the spoken word to written text is difficult. Pronunciations of words vary from person to person, and even vary from instance to instance. Most speech-to-text programs overcome this obstacle by first recognizing the voice of the speaker. Computerized voice recognition has been heavily researched since the 1980s, resulting in several patents [20], [21]. The voice is recognized by a frequency analysis of the higher frequency spectrum of the sounds. These higher frequencies are generated by the nature and use of the vocal cords and vary greatly between human speakers. After the voice is recognized, the programs look to databases that store information about words and pronunciations [22]. These databases are grown the more the program is used, so the program learns the human speaker. The location of the databases differs among different programs. One of the revolutionary components of Apples Siri is that the database is located on the cloud [23]. This allows phones to recognize voice commands without having to store a large amount of data.

Several open source options are available for speech recognition. With any of these programs, the input is spoken word and the output is a string of text. The next step is to assign meaning to those words. The challenge of this step is that humans naturally understand what is being said by the context. The same word can be interpreted different ways based on the context it is used. Sometimes the order of descriptive phrases is ambiguous. While humans are very adept at correctly sorting which parts of the sentence belong with which object, computers need explicit information to make decisions. Controlled languages attempt to translate text into a structured sentence.

*2) Controlled Natural Languages for Human to Robot Communication:* Natural human language is quite complex and many expressions have ambiguities. It will be even harder for a machine to understand since humans can not understand some of expressions by themselves. In order to resolve the language ambiguity problem and improve the efficiency of language understanding, Controlled Natural Languages (CNLs) are adopted in this project. CNLs are a subset of nature languages and use restricted vocabularies, grammars to eliminate the ambiguities in the languages. CNLs traditionally can be categorized into two groups: human oriented CNLs and machines oriented CNLs. Human oriented CNLs aim at producing better readability and comprehensibility documents for human to read. Machines oriented CNLs put the objective on improving the translatability, representation and processing of knowledges [24]. CNLs can be transferred into First-order logic which can be easily understood by computer systems.

There are many existing logic-based controlled natural languages such as Attempto Controlled English (ACE) [25], Computer Processable Language (CPL) [26], Processable EN-Glish (PENG) [27] to name a few. Compared with ACE, PENG uses a lighter weight but fully tractable subset of English. CPL interpreter resolves ambiguities by using heuristic rules. This is different from ACE which uses a small set of strict interpretation rules. It is also different from PENG which relies on predictive editor. Since ACE meets our need of transferring controlled nature language to first order logic and is already supported by various tools, we will use ACE as our method to make human language understandable to robot.

To make the robot understand human speech, a strong knowledge base is a crucial component. The knowledge base includes the tool category list, task list, object list, force and position profiles, off-nominal list, safety flag, and interface with other functional blocks.

As shown in Fig. 2(b), the communication from human to robot is in the first layer starting with human nature language, i.e. speech. In Section II-A1, we already discussed how to transfer speak to text which can be used for computer processing. The next step is to generate task logic using ACE. In some situations, a human will provide incomplete commands or wrong commands, e.g.,
*Human: Use saw to cut the I-beam.*
This is an incomplete command. After extracting the information and querying the information in knowledge base, it can be found there are two different saws, i.e. reciprocating saw, and band saw. Also the human has to show the location of the I-beam for the robot. Therefore, in this case, the knowledge base will return the incomplete information to the Robot to Human layer to ask human
*Robot: Which saw do you want me to use?*
*Robot: Where is the I-beam?*
Human will provide extra information by saying
*Human: Use Reciprocating saw.*
*Human: The I-beam is at P< x, y, z >.*

TABLE I
TASK LOGIC EXPRESSION

| Task logic | Description example |
|---|---|
| Normal command | Use Tool ($T_i$) to process($P_i$) on object($O_i$). |
| Object location | Object ($O_i$) is at <x,y,z>. |
| Off-nominal command | The tool ($T_i$) is not the right one, change tool ($T_j$) |

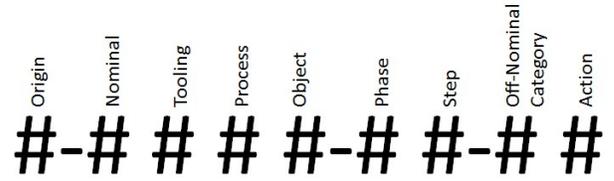

Fig. 3. Communication code.

In this paper, ACE serves as reasoning system to transfer language into first order logic format as

[*ACE Knowledge base*]: There is a reciprocating saw $T_1$. There is a band saw $T_2$. $T_1$ is used for cutting I-beam $O_1$. $T_2$ is not used for cutting I-beam $O_1$. $O_1$ is located at P< $x, y, z$ >.

Thus, after providing extra information for Task Logic block, the system will query ACE knowledge as

[*Query*]: *Cut the I-beam.*

Then the knowledge will return minimal subset of the knowledge to answer the query as

[*Answer*]:(There is a reciprocating saw $T_1$)∪ ($T_1$ is used for cutting I-beam $O_1$)∪( $O_1$ is located at P< $x, y, z$ >)

After that, the answer is generated into communication code as shown in Fig. 3 which will be discussed in detail in Section II-B. The communication code is an executable command to robot. The representation code for reciprocating saw is 1. Task code is 1. Object code is 2. Therefore, the communication code can thus be expressed by 3-111-23-41.

### B. Robot to Human Communication

While humans use five senses, the three dominant senses for most activities are sight, touch and hearing [28]. An advanced Cognitively Coupled Robot would be able to communicate on all three levels. As mentioned earlier, humans communicate most frequently and effectively using speech. It is through this channel that an entry-level cognitive robot will communicate. In order to translate from First Order Logic to human speech, several steps must be taken. First, the data must be organized and sorted. The robot is receiving information from various sources: the physical environment, the deductions of the robot task at hand, information about the database and tool learning engine, and information about the state of the communication process. These must be organized and prioritized. This task is accomplished by the Command Recognition sub-block in the Robot to Human block of Fig. 2(b). Each command is a series of numbers following an organized pattern.

The prioritizer applies weight to several categories to organize the delivery of the message. For instance, an off-nominal progress report from a task monitoring component will be communicated faster than a nominal progress report indicating that a tool learning routine was successfully entered into the robots data base.

The next sub-block queries a global robot database and returns a phase for each number of the code. As an example, 3-1111-23-41 might return (Task Monitor) (Off-Nominal) (Mitsubishi Reciprocating Saw) (Cut) (I-beam) (Cutting) (Initial Step) (Dull Blade) (Replace).

The final sub-block for the Robot to Human Communication block is to translate the text into human speech. The best approach for this step is to input the text into an online text-to-speech program. Several open-source options exist and could be used [29]. These programs will create an audio file that can be played over speakers.

### III. RECOLLECTION TOOLING INTERPRETATION FOR COGNITIVE TELEROBOTICS

#### A. Data Transformation

*1) Data Sorting:* A robotic arm is capable of carrying various of tools and sensors based on specific scenarios. The sensors, such as force/torque sensor, acoustic sensor, laser range finder and visual sensor, can be applied independently or merged to get comprehensive monitoring of the tooling status. Raw data, *e.g.* digital readings, sound frequency and visual images, are sent to the control system, however, they are hardly understandable by the tooling system before translated into characteristic information. Thus, in order to interpret sensor data, raw data is sorted by sensor type (see Figure 4). At the same time, some basic specifications such as sensor brand, type, model and reading unit of the sensor, are necessarily extracted from the Knowledge Base.

For the tooling task of using impact wrench to remove a conventional hex head fastener, basically three sensor are enough and applied to the robotic arm: force/torque sensor, acoustic sensor and a laser range finder. The acoustic sensor (microphone) reflects the process of removing the hex-head fastener by detecting the changes in frequency during the wrench's rotation. The force/torque sensor is used in fine positioning of the wrench, and the laser range finder is used for gross positioning. Similar to previous, the basic specifications

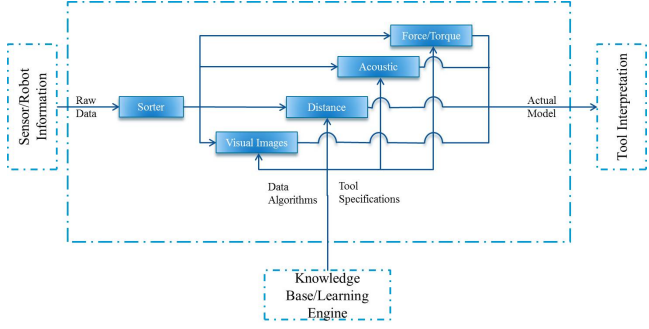

Fig. 4. The I/O of data interpretation block

is generated and responded by controlling the linear movement of the robotic arm.

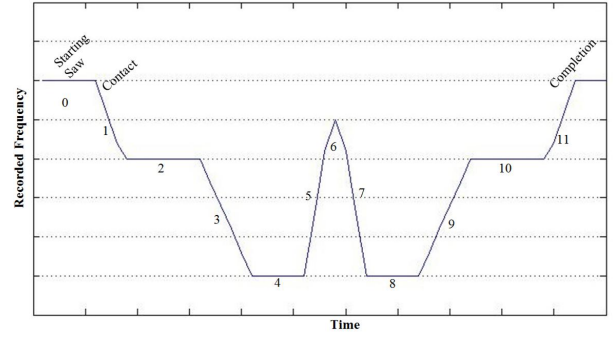

Fig. 5. Frequency profile for cutting a 45° inclined I-Beam with inset step numbers

are extracted from the Knowledge Base to be used in the profile generation.

*2) Tooling Model Profile Generating:* Following the sorted sensor information, the data algorithms and tool specifications are pulled from the Knowledge Base, which can interpret raw sensor reading and generate corresponding profiles. The data algorithm is applied and also based on specific tool characteristic, to generate profiles for force/torque, sound frequency, distance or visual images. The tooling profiles can be merged into one tooling model or updated individually, depending on the requests from Tool Interpretation block. Generated profiles make up the actual model characterization and the model is then sent to the Tool Interpretation for comparison with the theoretical model from Control System, in order to determine the state of the process.

*a) Reciprocating saw cutting an I-beam:* The Knowledge Base stores the specifications of I-beam with different dimensions and the profiles of force/torque, sound frequency and distance during the cutting. Visual profile can also be an option for monitoring when off-nominal case happens. Based on the reciprocating saw type and the dimension of target I-beam, detailed specification and dimensions are loaded from the knowledge base. Before applying the loaded profiles for the actual model, some assumptions for the frequency profile are considered:

1) The saw's frequency is inversely proportional to the cutting area.
2) The saw progresses at a constant speed.
3) The saw blade has negligible thickness.
4) Neglect the round corner of the I-beam, assume it has squared corners.

Once the assumptions and cautions are clear, the profiles are started to generate. The cut operation will be broken into three maneuvers: Approach, Cut and Retract. During the Approach routine, the saw will begin vertically and slowly rotate about a fixed point about 20 inches from the tip of the saw. This phase will end when the saw blade comes in contact with the I-beam. The saw will then rotate about the contact point until it is horizontal. This process will also identify the top right corner of the flange. The saw will raise about 0.25 inch and move horizontally so that the tip of the I-beam is about 4.5 inches from the base of the blade. The distance control is monitored through the laser range finder, thus the range profile

The Cut routine will begin with turning the saw on, generating a starting frequency level (see Figure 5). According to the I-beam, during the Cut routine, the saw will descend vertically and retract horizontally until the $5^{th}$ profile is recognized. Then the saw will descend vertically and extend horizontally until the $7^{th}$ profile is recognized. The final point involves descending vertically and retracting horizontally until the I-beam is fully cut. Corresponding distance for the saw has travelled during cutting is shown in Table II. The sound frequency from the saw blade when touching an object is inversely proportional to the cutting area, such as when the blade is dull, the frequency increases, thus the sound frequency profile is used as the major method for monitoring the process. The force/torque sensor can be used as a backup plan to assist the microphone, once the microphone goes wrong. The force profile is shown in Figure 6.

TABLE II
I-BEAM DISTANCE PROFILE

| Profile point | Distance saw traveled | Cutting area length |
|---|---|---|
| 1 | $\frac{t_f}{\sqrt{2}}$ | $\sqrt{2}t_f$ |
| 2 | $\frac{t_f}{\sqrt{2}} + \frac{b_f - t_w}{2\sqrt{2}}$ | $\sqrt{2}t_f$ |
| 3 | $\frac{t_f}{\sqrt{2}} + \frac{b_f}{2\sqrt{2}}$ | $\sqrt{2}t_w + \sqrt{2}t_f$ |
| 4 | $\frac{b_f}{\sqrt{2}}$ | $\sqrt{2}t_w + \sqrt{2}t_f$ |
| 5 | $\frac{t_f}{\sqrt{2}} + \frac{b_f}{\sqrt{2}}$ | $\sqrt{2}t_w$ |
| 6 | $\frac{d - t_f}{\sqrt{2}}$ | $\sqrt{2}t_w$ |
| 7 | $\frac{d}{\sqrt{2}}$ | $\sqrt{2}t_w + \sqrt{2}t_f$ |
| 8 | $\frac{d}{\sqrt{2}} + \frac{b_f}{2\sqrt{2}} - \frac{t_f}{\sqrt{2}}$ | $\sqrt{2}t_w + \sqrt{2}t_f$ |
| 9 | $\frac{d}{\sqrt{2}} + \frac{b_f}{2\sqrt{2}} - \frac{t_f}{\sqrt{2}} + \frac{t_w}{\sqrt{2}}$ | $\sqrt{2}t_f$ |
| 10 | $\frac{d}{\sqrt{2}} + \frac{b_f}{\sqrt{2}} - \frac{t_f}{\sqrt{2}}$ | $\sqrt{2}t_f$ |
| 11 | $\frac{d}{\sqrt{2}} + \frac{b_f}{\sqrt{2}}$ | 0 |

The final routine moves the saw to the original position while avoiding the I-beam. This is a controlled horizontal retraction, then rotation to vertical, followed by rising to the original position. The profile generated in this routine is to lead robotic arm to move back, and the distance is detected by laser range finder.

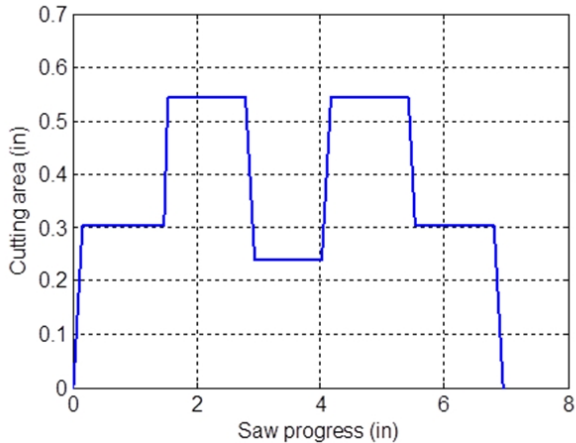

Fig. 6. I-beam force profile

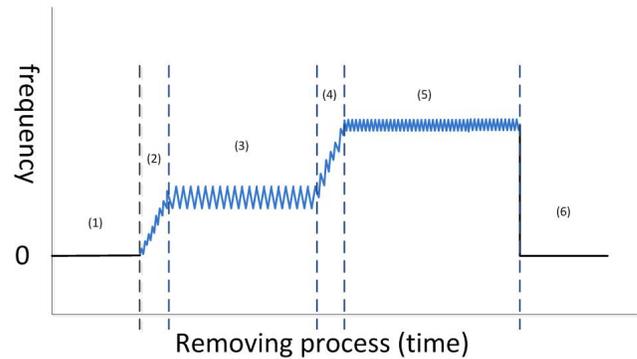

Fig. 7. bolt head removing sound frequency profile

The visual sensor is used as an auxiliary tool for monitoring the tooling process. Image processing algorithm is loaded from knowledge base and exerted to sorted visual images. Extracted visual profiles can also be applied for compensating model characterization if needed.

*b) Impact wrench removing a conventional hex head fastener:* In this task, the frequency and distance profiles are generated as well as force/torque thresholds loaded. Given the sorted raw sensor data, i.e. frequency and distance, tool specifications are pulled from the Knowledge Base. The tool task is broken into three maneuvers: Approach, Remove and Retract. During the approach routine, the manipulator brings the wrench to find and approach to touch the bolt. The force/torque profiles will not generated during this step since no force is applied, as in Figure 8 step (1) shows. The distance profiles are generated by recording sorted range finder data. Once the wrench is brought over the bolt head, the manipulator moves forward a certain distance to cover the bolt head. The manipulator can give a brief one second burst of the wrench to determine if there is resistance present, indicating whether the tool is properly centered on the bolt. If the distance doesn't match the fastener specification in knowledge base or the burst has little resistance, the manipulator adds a 10 to 15 degree twist about the wrench head to settle the wrench closely cover the bolt head. Once the force/torque data generated, this alignment process is finished.

The removing routine starts once the sound frequency profile generated. With the help of sound frequency profile from microphone, the unbolting process is becoming easy and efficient. As Figure 7 shows, in step (1), the wrench closely covers the bolt head. Step (2) represents the starting process, until its steady working routine. The sound frequency of step (3) is relatively low but with higher amplitude. The sound frequency can be used to replace the tentative burst in previous routine without microphone. Thus in this routine, the unbolting process does not stop until the sound frequency reaches it free-air spinning frequency, such as the part (5) in the figure, which has lower amplitude but higher frequency. If the sound frequency is detected to last for a certain time, then the removing routine is considered to be finished and the wrench stop working as in figure step (6) shows. Similarly, the force/torque sensor is used as backup profile for obtaining the process when the microphone does not work properly. The force profile is shown in Figure 8, step (2) part.

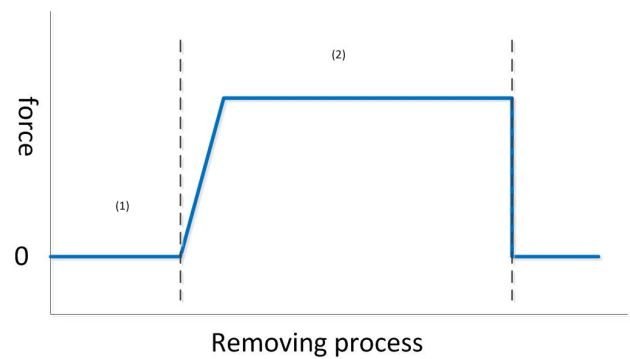

Fig. 8. bolt head removing force profile

Once the removing process is identified to be finished by no more resistance detected, or the wrench is running in steady free-air spinning, the manipulator retract the tool to its original place. The distance profile is generated for later Knowledge Base update. If applying sound frequency, the determination of the initiating of retract routine becomes much simpler and intuitive. However, if to estimate the initiating of retract routine without a microphone, a translational force is set. Applying translational force as a threshold is a means of testing whether or not the bolt has been sufficiently unscrewed such that it is no longer within its hole. This threshold allows for a quick test without requiring the tool to be removed from the bolt, necessitating the need for realignment after checking.

*B. Tool Interpretation*

As the profiles are coming in from the Data Interpretation block, they are immediately compared to the theoretical outcomes sent from the Control System/Task Planning. This allows for a determination of whether or not the present operation is currently going nominally or has gone off-nominal. This determination comes by means of an error percentage.

Should the percent error fall at four percent or under in difference (providing room for general fluctuations within the operation), the operation would be regarded as still within nominal operating range. Beyond a four percent variation between the actual outcome and theoretical outcome would be construed as off-nominal. The block is shown as in Figure 9.

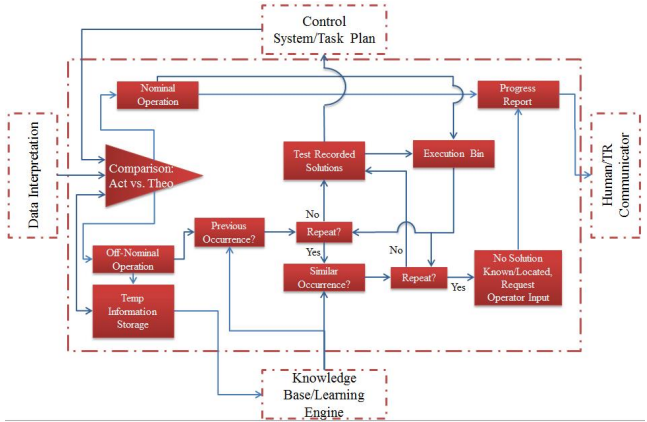

Fig. 9. The I/O of tool interpretation block

*1) Nominal Operations:* If it is determined that the operation is performing nominally, a progress report is sent on to the Communication section, and the operation continues forward. A secondary command is sent to the Execution Bin clearing it of any stored information (explained shortly).

*2) Off-Nominal Operations:* Once the operation is judged to be going off-nominal, a copy of the off-nominal operation data is stored in a temporary bin called Temp Information Storage while another is sent on to begin a process of "Recollection Testing".

*a) Recollection Testing:* The portion sent on is compared to the Knowledge Base's stored information of previous off-nominal operations. The particular past operation with the lowest percent difference to the present operation is considered a previous occurrence and its correction procedure script is then compared against the procedures stored in the Execution Bin to determine if this specific procedure has already been attempted. This particular comparison step is critical in preventing the telerobot from becoming trapped within a continual loop of "Idiocy" where it would continuously attempt the same (ineffective) procedure over and over. If the procedure has not already been attempted, it will pull up the attached correction procedure script and send it on to the robot's Control System/Task Plan to be executed.

*b) Temp Information Storage:* This particular part of the Interpretation process could be inferred as the actual "learning" portion. The Temp Information Storage receives and stores a copy of the data that was judged off-nominal by the comparison of the theoretical outcome versus the actual outcome. Throughout the entire Recollection Testing phase of the off-nominal process, the information within the storage bin is not used. However, it comes back into effect once the comparison between the theoretical outcome and the actual outcome determines a nominal operation.

Once the present operation is judged to be nominal, a brief scan of the Temp Information Storage bin is performed to determine if any information is currently being stored within it. If there is no information stored (operation never went off-nominal), nothing else occurs. On the other hand, if information is found to be within the Temp Information Storage bin, a copy of the procedure script just executed (received from the Control System/Task Plan with the theoretical outcome data) is sent to the storage bin. Receiving the procedure script, the Temp Information Storage bin attaches the two pieces of information together as a single file and sends it to the Knowledge Base for future occurrence comparisons.

## IV. Tooling Knowledge Base with Learning Capability for Cognitive Telerobots

### A. Preliminary Knowledge

*1) Knowledge Base:* A variety of knowledge bases have been proposed by the computer science society. The data structure we used is a C++ class based codes. All information needed for the cognitive telerobot will be stored in different classes. C language is widely used and we plan to use them on a Labview program where interface to various of hardware is also available.

*2) Sensors Used in Teleoperation:* There are several types of sensors used in the system: force/torque sensors, acoustic sensors and position/velocity sensors. The force/torque sensors are usually strain gauge based. They are resistor put on parts which their resistance change when the parts deform. Amplifier circuits have to be used on them to amplify the signal. There are many types of position/velocity sensors: encoder, potentiometer and resolver. Encoders are digital equipment which can be very accurate. Potentiometers are very robust analogue equipment with very simple interface. Resolvers are complicated but the most accurate position sensor available. The acoustic sensor used is a high resolution microphone which turns sound wave into time domain voltage signals.

*3) Reinforcement Learning:* For the tool learning engine, a general solution using reinforcement learning (RL) algorithm [30], [31], [19] was employed in this paper. RL is concerned with how an agent should take actions in an environment so as to maximize the value of cumulative reward. The environment is typically formulated as a Markov decision process (MDP) [32], RL is usually used to solve this problem in the context of dynamic programming techniques.

The basic reinforcement learning model applied to MDP mainly consists of three elements, including a set of environment state $S$, a set of actions $A$ and a real valued rewards function $R(s)$. A robot interacts with the environment continuously. At each time $t$, the key idea in RL is to update and store a Q-factor matrix for each state-action pair in the system. Thus, $Q(i, a)$ will denote the Q-factor in term of state $i$ and action $a$. At the beginning, the values in Q-factor matrix are initialized to suitable numbers. Then the system is trained using the algorithm. When a state is visited, an action is selected and the system is allowed to transfer to the next state. The immediate reward that is generated in the transition is recorded as the feedback. The feedback is used to update the Q-factor for the action selected in the previous state. Generally speaking, if the feedback is good, the Q-factor of that particular action and the state in which the action was selected is increased (rewarded). Otherwise, the value of Q-factor is reduced as a punishment.

### B. Data Structure of Knowledge Base

*1) Tooling Knowledge Base:* The tooling knowledge base is a tool-centric dynamic database base. The data structure that we adopted for list is a type of 'class'. Each instance of that 'class' is corresponding to a specific model of a type of tools. Each instance has the tool model number, functionality index, task indices, force/torque profile and safety flags. These data cover all aspects of the tools. Several look-up tables regarding the actual meaning of the indices are predefined and updated whenever necessary.

*a) Categories List:* The tools are categorized into several major classes. The most commonly used ones in our experiments are the reciprocating saw and the nut runner. Tools in one category usually shares the same type of mechanical and electrical structures as well as the type of functionality.

*b) Task List:* The tasks are all basic descriptions about one general type of job. Some general information about this kind of task will be included.

*c) Off-nominal List:* This is one of the most difficult tools to handle among all the lists. Most of the learning and experiencing process is about updating and reviewing this list. The list includes information about the off-nominal cases that the system have encountered or the human interpretation has informed about. Then a solution will be proposed to deal with this Off-nominal situation.

*d) Force and Position Profiles:* Force and position profiles have been used to determine the status of the current operation. Previous analysis considers the profile as a curve made up of data points, while in our analysis, a parametric style data will be extracted from the raw data to save storage space. For example, the force profile of cutting a pipe. There are two peaks on the force curve, which corresponds to the two peaks of the cutting area during the process. And the time spent is related to the actual size of the pipe.

*e) Interface with other Functional blocks:* This block interfaces with the 'Tool Operational Interpretation', 'Tool Learning Engine' and 'Tool Operational Characterization Model'. The 'Tool Operational Interpretation' provided actual measurement compared with the task plan. 'Tool Operational Characterization Model' is a model extracted from actual sensor data. 'Tool Learning Engine' provides the update from the human interpretation.

*2) Sensor Knowledge Base:* There are three major types of sensors: position/velocity sensors, force/torque sensors and the acoustic sensor involved in this paper. Some sample interpretation is shown in Tab. III.

*a) Position/Velocity Sensing:* A set of position and velocity data collected earlier is fed into the computer. The computer first smooths the data using a band pass filter, and then the features needed to determine the status is extracted. These features are compared with a corresponding entry in the knowledge base.

*b) Force/Torque Sensors:* Force/torque profile from the procedure of traditional teleoperation is sent to the control computer of cognitive teleoperation to be tested. The dimension, rigidity, material and other properties are also sent to the computer. The computer pulls out the correct entry from the knowledge base table and makes a comparison.

*c) Acoustic Sensor:* Fourier transform is applied to the data collected from acoustic sensor to generate a frequency spectrum. When the saw is in contact the the I-beam, its speed is substantially reduced because of the friction caused by the cutting process. This information can be used to generate a frequency spectrum to diagnose the status of the system.

*3) Script Converter:* A simulated task command coded using integers are sent to the knowledge base, the computer program looks into the table and find the corresponding scripts to the 'Task Planner'. During the test, every possible combination will be tested. If there is any that is not feasible, the computer will generate a error code.

### C. Tool Learning Engine

In this section, tool learning engine was used to update the information about current subtask in the knowledge base, including the tool signature, tool position, maximum of force and flags about off-nominal operations. Here we consider two tasks: 1) cutting an I-beam using a reciprocating saw, 2) removing a conventional hex head fastener with an impact wrench. For the first task, the target is an I-beam placed at a 45 degree angle with respect to the saw. The cutting blade is perpendicular to beam cross-section. The desired applied forces should be in an acceptable range so that the task is implemented efficiently, and the tool, manipulator and other components are not damaged. In later task, the object is a conventional hex head fastener. Similarly, the applied forces should be acceptable so that the task can be performed efficiently without any damage.

Learning modes are involved in the block of tool learning engine. Learning from experience is the triggered in the whole task. It automatically updates the information relative current work according to the state of environment. The reinforcement learning algorithm is employed in this mode. For the second mode, the telerobot can learning from the rule command from human operator. It can help the learning of robot by enriching the knowledge base. In the following sections, we take I-beam cutting using a reciprocating saw for example to discuss two modes in details.

*1) Learning by Experience:* In this section, to enable the continuous behaviours learning, RL is employed to learn at the global and local level. Global level is used to learn the whole processing of task in term of the state of environment, while the local learning is performed for the specific subtask.

*a) Global Learning:* For simplicity, only two states are considered in this model: nominal and off-nominal cases. Both cases are described by the state of environment, such as tool position, attributes of the target, force signature, maximum of force and statistics of task (e.g. running time, duration of blade and sharpness of blade). In fact, the states can be divided further according to the environment. A robot can choose possible actions, including *Approach*, *Cut*, *Retract* and *Replace the blade*. We assume the actions of *Approach* and *Retract* always work in the nominal case. Improper action of *Cut* can lead to off-nominal operations, e.g. blade is broken or dull. The off-nominal state can return to nominal one by implementing the action of *Replace the blade*.

TABLE III
EXAMPLE OF SENSOR DATA INTERPRETATION

| Sensor | Data | Status |
|---|---|---|
| Acoustic sensor | Too low | Insufficient power/Pushing too hard/Dull blade |
|  | Normal | Normal |
|  | Too high | Broken blade/Out of contact |
| Position sensor | Error greater than desired | Desired trajectory too agressive |
|  |  | Collision |
|  |  | Needs lubricant/maintenance |

The reward function $R(s)$ is determined by the execution performance of the task. if the task is completed without off-nominal events, the reward is set in the range of $(50, 100]$ depending on the efficiency (running time). The Higher the efficiency is, the more rewards added. if the robot finishes the task with off-nominal, the rewards in $[0, 50]$ is assigned. The worst case occurs when the task is not done with off-nominal operations, and the penalty value is set to $-50$. Obviously, the pair of state and action with less rewards makes less contribution for the whole task. The details are shown in Table IV.

In this paper Q-learning was employed [33]. The performance can be optimized by the rewarding process without being supervised. By the experience, the robot can figure out which action is suitable according to the state of environment. To improve the performance of robot operation, Q-learning optimizes an action-reward function that estimate the efficiency of an action in a certain state. This action-reward function is learned by exploring the state-space following a certain action. To implement Q-learning in this paper, two matrices $R$ and $Q$ are introduced. $R$ stores the state dependent rewards. For instance, if the task is not done in the off-nominal state, a penalty of $-50$ will be assigned to $R$ according to the corresponding state. Matrix $Q$ represents the total knowledge so far. It is related to the state and action, and usually initialized as a zero matrix.

In the Q-learning technique, the main problem is how to construct and optimize the matrix $Q$. In this paper, $Q$ is learnt by the Eq. 1.

TABLE IV
ELEMENTS IN REINFORCEMENT LEARNING MODEL.

| States | Actions | Rewards |
|---|---|---|
| Nominal and off-nominal (tool position, cutting sound, sharpness of blade, target attributes, force signature, maximum of force, running time of task) | Approach, Cutting, Retract, Replace the I-beam | 1. Cut off the blade without off-nominal events: Reward (50 100] 2. Cut off the I-beam with off-nominal events: Reward [0 50] 3. The I-beam is not cut off with off-nominal events: Penalty -50 |

$$\hat{Q}(s,a) = r(s,a) + \gamma \max_{a'} Q(s',a') \quad (1)$$

where $\hat{Q}(s,a)$ is the new value after learning, $\gamma$ is the learning rate in $[0, 1]$. It determines the importance of new knowledge. A value of 0 for $\alpha$ means anything is not learnt by the robot, the knowledge based is not growing, while the newest knowledge will be accepted with a value of 1. $s'$ and $a'$ are new state and possible actions, respectively.

In this paper, matrix $Q$ is considered as a look-up table. After $Q$ is obtained with the relatively complete knowledge, given a task, we can estimate the state for each subtask by searching in the matrix, including the tool signature, maximum of force and so on.

*b) Local Learning:* Here we considers a common scenario in construction: cutting a standard I-beam with a reciprocating saw. There are several things needed to be chosen before the operation and several parameters that needed to be chosen during the operation. For example: the force applied on the reciprocating saw is controlled by the robot and it can be considered as a "action" of the system, and the state of the system can be: good progress, slow, fast, broken, etc. A Q matrix could be initialized and updated during the process of the cutting. The reward function can be designed from basic engineering intuition: negative reward for broken blade, small reward for slow progress and great reward for fast progress. So this learning process is a emulating of the human learning process.

Most of the time when this algorithm is used, the initial value of Q matrix is a random number matrix, which means the robot have no prior knowledge about the task and tool. In our case, we can make the robot has some predefined knowledge, that is to say, assign some initial value to the Q matrix. That way, the system can learn a lot faster.

*2) Learning from Operator:* Sometimes the knowledge between the robot and operator is not consistent, the operator should send the rule command to teach the robot. The new skill will help robot adapt the new situation. The rule command is sent using the voice input [34]. The nature language is translated into the first-order logic [35]. However, the robot is often instructed without fully specified command, which means it does not include all the necessary information to derive a sequence of actions for the robot. Hence uncertainty exists in the execution of the command.

To address this issue, a robot need to infer information that is missing in commands according to the knowledge stored in the matrix $Q$ obtained in Section IV-C1. A procedure of knowledge fusion is performed by coupling the information from command and existing knowledge, so that a complete knowledge is produced, and added into *Q* to update the knowledge base.

## V. Conclusion

Robotic cognition will continue to be a driving component of telerobots as we place them in increasingly unstructured and unknown environments. By following the structures outlined in this paper, humans will be able to interact with robots on a cognitive, rather than sensory, level. This cognitive coupling will allow robots to make better decisions about tasks at hand and to learn new approaches and methods faster. The success of robot-human cognitive coupling lies in the understanding of human and robot thought. A translation must occur between human thought, which is represented by words and phrases that have associative meaning, and digital thought, which is organized by first order logic. Developments in Controlled Natural Languages, Text-to-Speech algorithms and Speech-to-Text acoustic models have provided the infrastructure needed to begin work on an entry level, speech-driven cognitively coupled robot.

A tool-centric knowledge base with learning capability was presented as our third goal. The knowledge base was organized in a compact, fast and economic way. The learning capability allows the continuous update of knowledge based to deal with the uncertainty of the dynamic environment and unspecific operator's command. The reinforcement learning was employed to learn by experience and update the knowledge base at the global and local levels. Also, the knowledge base was enriched by incorporating operator's cognition. Therefore, the cognitive level of robot grows along with the underlying learning. The robot with these knowledge from learning can work on different tasks with off-the-shelf tools.